\documentclass[]{spie}  

\usepackage{amsmath,amsfonts,amssymb}
\usepackage{graphicx}
\usepackage[colorlinks=true, allcolors=blue]{hyperref}

\usepackage{times}
\usepackage{epsfig}
\usepackage{graphicx}
\usepackage{amsmath}
\usepackage{amssymb}
\usepackage{booktabs}

\usepackage{tikz}
\usepackage{comment}
\usepackage{color}

\usepackage{multirow}
\usepackage{lscape}
\usepackage{longtable}

\usepackage{epsfig}
\usepackage{booktabs}
\usepackage{adjustbox}
\usepackage{multirow}
\usepackage{pifont}%

\usepackage{color, colortbl}
\usepackage{caption}
\usepackage{algpseudocode}
\usepackage{subcaption}
\usepackage{soul}
\definecolor{darkmagenta}{RGB}{127,0,127}
\definecolor{LightCyan}{rgb}{0.88,1,1}
\definecolor{whitesmoke}{rgb}{0.96, 0.96, 0.96}

\title{In-situ Water quality monitoring in Oil and Gas operations}

\author[a]{Satish Kumar}
\author[b]{Rui Kou}
\author[b]{Henry Hill}
\author[b]{Jake Lempges}
\author[b]{Eric Qian}
\author[b]{Vikram Jayaram}
\affil[a]{University of California, Santa Barbara}
\affil[b]{Pioneer Natural Resources,  Irving, Texas}

\authorinfo{Further author information: (Send correspondence to Satish Kumar, Rui Kou, Vikram Jayaram)\\Satish Kumar: E-mail: satishkumar@ucsb.edu, \\  Rui Kou: E-mail: kourui.mun@gmail.com, \\Vikram Jayaram: Vikram.Jayaram@pxd.com}

\begin{document} 
\maketitle

\begin{abstract}
From agriculture to mining, to energy, surface water quality monitoring is an essential task. As oil and gas operators work to reduce the consumption of freshwater, it is increasingly important to actively manage fresh and non-fresh water resources over the long term. For large-scale monitoring, manual sampling at many sites has become too time-consuming and unsustainable, given the sheer number of dispersed ponds, small lakes, playas, and wetlands over a large area. Therefore, satellite-based environmental monitoring presents great potential. Many existing satellite-based monitoring studies utilize index-based methods to monitor large water bodies such as rivers and oceans. However, these existing methods fail when monitoring small ponds--the reflectance signal received from small water bodies is too weak to detect. To address this challenge, we propose a new Water Quality Enhanced Index (WQEI) Model, which is designed to enable users to determine contamination levels in water bodies with weak reflectance patterns. Our results show that 1) WQEI is a good indicator of water turbidity validated with 1200 water samples measured in the laboratory, and 2) by applying our method to commonly available satellite data (e.g. LandSat8), one can achieve high accuracy water quality monitoring efficiently in large regions. This provides a tool for operators to optimize the quality of water stored within surface storage ponds and increasing the readiness and availability of non-fresh water. The code-base is publicly available at Github: \url{https://github.com/satish1901/In-situ-Water-quality-monitoring-in-Oil-and-Gas-operations}

\end{abstract}
\vspace{-0.45cm}
\section{Introduction}

Water is one of the most abundant and most essential resources on earth.  Not only does water sustain all life on Earth, but it is also requisite to industrial processes such as fabrication, washing, cooling, and fuel generation~\cite{usgs_water,un_gwc}.  
Particularly in the oilfield, the average frac job uses 4 million gallons of water, and availability becomes a key logistical issue in high activity basins such as the Permian~\cite{frac_job}. The industry addresses this necessity with efficient storage of water, particularly in networks of open body ponds down to the frac site~\cite{pnr_report}.  However, new challenges arise from even this, such as monitoring, sourcing, transportation, and treatment. 


The monitoring and maintenance of these water bodies is crucial. Monitoring is usually done manually: water samples are collected from different points in the pond (as impurities can vary spatially) and sent off to a lab to be tested. Tests include checking for types of impurities, sediment concentration, algae growth, turbidity of the water, and chloride concentration changes over time. 
The amount of manual field work needed to collect samples from multiple points and multiple water bodies across a large distance is almost infeasible. Due to the logistical challenges testing presents, sampling and testing is conducted at extremely infrequent intervals. To make matters worse, tests done on water samples do not produce in-situ results.

To overcome these major limitations, we propose our novel Water Quality Enhanced Index ($WQEI$) to detect the turbidity and salinity in water ponds, open tanks, water storage ponds for irrigation, playas, etc using multispectral satellite images ~\cite{landsat8,sabins1999remote}. Remote Sensing has been used in multiple domains~\cite{kumar2020deep, kumar2022guided, jayaram2004detection}, often to create spectral indexes.
Spectral indexes are combinations of spectral reflectance from two or more wavelengths that indicate the relative abundance of features of interest~\cite{AllIndexes, indexes, harrisIndexes}. Vegetation indices are the most popular of these, derived using the reflectance properties of vegetation~\cite{8746029,anderegg2020spectral,vegIndexes,harrisIndexes,vegIndexCcpo}. The most popular vegetation index is NDVI (Normalized, Difference Vegetation Index)~\cite{8746029,anderegg2020spectral}.

The question of water quality detection using satellite imagery has seen much scientific inquiry already~\cite{usali2010use,bonansea2019using,hua2017land,liu2003quantification,chebud2012water,seyhan1986application,griffith2002geographic,ritchie2003remote,schaeffer2013barriers}.However most of this work detects water quality in large water bodies with substantial depth($\geq \sim100$ft) and size($\geq100$ acres). The reflectance signals recorded in such cases have a high Signal-to-Noise Ratio (SNR). There is minimal interference from surrounding confounding elements like the bottom of the water body and pixels with low SNR.
Small-sized water bodies present unique challenges and prevent the translation of existing methods directly to monitor small ponds, shallow lakes, storage tanks, playas, etc. The small size ($\sim 2$ acres) causes the pixels representing surface reflectance in the satellite image to be very noisy due to interference from surroundings(soil, rocks, bushes, and other potentially confusing elements on the ground). Shallowness($\sim 30$ ft) of the water body also adds noise to reflectance values: the earth below has a different surface reflectance than the water above. In this study, we focus on the following areas: 
\begin{itemize}
    \item We propose a novel Water Quality Enhanced Index ($WQEI$) for detecting the turbidity, and salinity in very small and shallow water tanks, ponds, playas using satellite images.
    \item The proposed index improves the SNR of the weak signal received from the ground. It is robust to changing conditions in the surroundings. e.g. changes in the vegetation type, in soil salinity, etc.
    \item We verified the output of our estimation index against lab tests done on 49 water ponds for hydraulic fracturing purposes. This data was collected over a 2-year time period for each pond at variable frequency, totaling $\sim 1200$ samples.
    \item We tested our index on multi-spectral data from two different sources of satellite imagery: commercial satellite (private) and LandSat8 (public).
    \item We developed an end-to-end pipeline that pulls data from LandSat8 periodically every week. This allows us to monitor any water body anywhere on earth.
    \item We also developed and trained a neural network to check the effectiveness of machine learning models for such problems.
    
\end{itemize}

Overall we developed a cost-effective and efficient technique for monitoring small water bodies' turbidity and salinity. These small sources of water can thus become usable to prevent wildlife from consuming polluted water, and in the restoration of natural playas and a healthier ecosystem.




\section{Related Works}
Our method draws from numerous areas of remote sensing, utility of spectral signatures, and information capture from the landscape. In this section, we discuss the key relevant areas of existing works that motivated our work and contributions. 
\newline
\textbf{Deterministic and Semi-empirical methods:} There have been many studies done on detecting different types of impurities in water~\cite{bonansea2019using,hua2017land,liu2003quantification,chebud2012water,deekshatulu1981application,seyhan1986application,griffith2002geographic,ritchie2003remote,schaeffer2013barriers,usali2010use}, these studies are often designed for a specific water body in a specific region.
For example, ~\cite{usali2010use} took the data from a 1978 Malaysian water monitoring program and developed estimation methods for suspended matter~\cite{dekker2002analytical}, phytoplankton growth~\cite{han2005estimating} and turbidity~\cite{wang2006applications}, using multi-spectral data from LandSat5 satellite. That monitoring program had created data for only two similar rivers within Malaysia, which greatly limits its generalizability. 
~\cite{ritchie2003remote} used multi-spectral data from LandSat5 (TM) to analyze the presence of suspended sediments and chlorophyll in Lake Chicot, Arkansas.~\cite{griffith2002geographic} took a different approach and proposed the importance of detecting landscape features while assessing the impurities in a water body. ~\cite{griffith2002geographic} analysed the relationship of landscape pattern ecological processes. 
For example, environmental attributes and processes like water quality, nutrient flow, and population dynamics are correlated with landscape spatial patterns using different indexes. 
~\cite{deekshatulu1981application} studied the attenuation coefficient of water by mixing different types of impurities in it and analyzed the relationship between spectral reflectance and different level of depth of water. This study was conducted on 432 acres of Loosdrecht lakes in the Netherlands. 
\newline
\textbf{Regression and Neural Networks based methods:} Regression analysis is the most common data-driven approach in remote sensing.~\cite{zhang2003water} mentioned the requirement of comparing multiple approaches e.g. deterministic, semi-empirical, and empirical, data-driven regression analysis methods when computing the water quality index.~\cite{ritchie2003remote,hu2009novel,zhang2002application,zhang2003water,hu2018introduction,sawaya2003extending} mentioned different data-driven approaches.~\cite{hu2009novel, zhang2003water} for the first time mentioned that water quality measurement methods were dependent on the water body classification (i.e, lake, pond, playa, tank, etc) and water depth that needs multiple bands information.~\cite{zhang2002application,zhang2003water} calibrated empirical relation between different bands in LandSat8-OLI/TRS imagery to detect chlorophyll present in the water body.~\cite{ritchie2003remote} evaluated the possibility of a nonlinear relationship model by checking the water bodies in  Arkansas and Mississippi, USA.~\cite{ritchie2003remote} claim was supported by~\cite{sawaya2003extending} of using regression analysis.~\cite{sawaya2003extending} evaluated the combination of multiple bands and proved that no single band combination has uniqueness, thereby implying the use of information from all the bands to make any kind of decision.

\begin{figure*}[t]
\begin{center}
\includegraphics[width=\linewidth]{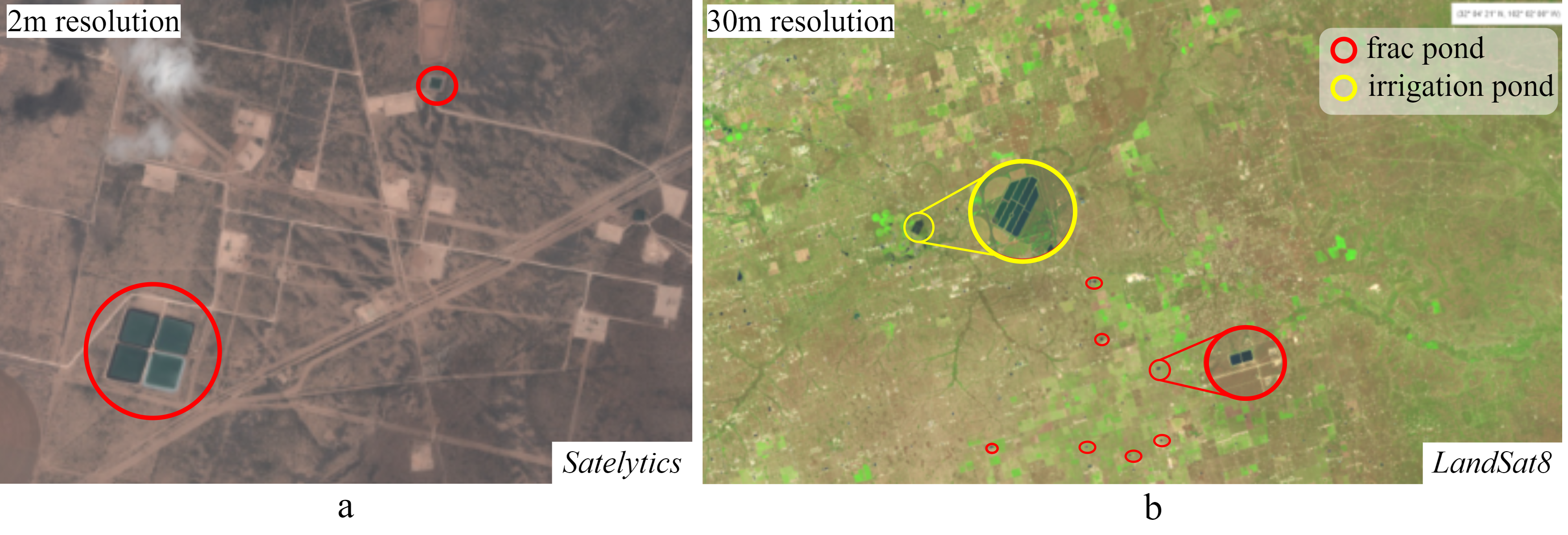}
\end{center}
\vspace{-0.45cm}
   \caption{Image ``a"shows data sample from private source at 2m per pixel resolution, the marked regions are frac ponds. Image ``b" shows data from LandSat8 satellite at 30m per pixel resolution, it shows different types of water storage ponds. It can be seen that at 30m resolution the ponds contribute only few pixel in the image that makes the problem more challenging to detect $WQEI$.}
\label{fig:intro }
\end{figure*}

Neural Network approaches have been proven to be better than traditional methods~\cite{jensen1996introductory, kumar2022locl}. 
The usefulness of the neural networks was tested by~\cite{sudheer2006lake} to compute the water quality index.~\cite{sudheer2006lake}'s use case proved that neural network captures all the vital, non-computable by regression methods relations like atmospheric disturbance, non-ideal contextual uncertainties, etc. The McCulloch and Pitt model~\cite{hu2018introduction} is the widely used network. It consists of a simple multi-layer neural network with convolution layers and non-linear activation functions in between batch normalization. 
With all the benefits and flexibility of neural networks, the major limitation is that the convergence of neural networks is very difficult with less amount of data. The curse of dimensionality if a huge issue in neural networks. Since the multispectral data is very high in dimension with very small numbers of pixels of interest. It needs a large amount of clean training data and big compute power(GPUs) to train a neural network. This was one of the motivations for our work of selecting simple nonlinear regression methods for creating the water quality index explained in detail in the next section.

\section{Approach}
In this section, we describe the technical approach for our Water Quality Enhanced Index (WQEI) in detail, we will also talk about the study area, details about the dataset used, and satellite information.

\subsection{Study Area} The area selected for the development, analysis, and evaluation of $WQEI$ in Midland, Texas, USA. This region uniquely contains a large number of frac ponds closer to the oil drilling sites along with a large number of playas in the same region. Right next to the Midland region, there are many ponds for irrigation purposes in Stanton, Texas, USA. Along with these, there are many small natural water bodies in the Permian basin, Texas region, that appears during the monsoon season and disappear over the year as summer progresses. 
This region uniquely contains water bodies of different sizes, shapes, and depths. Another major benefit of these sites is that they have different use cases (i.e. fracing job in oil well drilling, irritation purpose, industrial use, drinking purposes, etc). That adds a different types of sediments and impurities, this made our $WQEI$ more robust and generic to be used for different types of water bodies. Figure~\ref{fig:sample_bodies} shows few samples of different types of water bodies tested in our case. The water stored in frac ponds is built to store fresh water or flowback water from the well, or a mixture during the course of well-site development. It is most important to monitor frac ponds as the water which flows back from wells may contain some dangerous substances, and should be monitored for those according to federal, state, and local laws~\cite{frac_ponds}. Usually this water becomes unusable overtime, and it adds a huge economic burden on the oil \& gas companies and damage to environment.
Some of the irrigation ponds are retention ponds that capture stormwater. Texas alone has more than one million ponds and small farming lakes~\cite{irrigation_ponds}. This is an extremely large number to be monitored manually, which is where remote sensing is the most effective. 
The variety of sizes, shapes, depths uses and seasonalities make this region an ideal environment for a monitoring study. 

\subsection{Dataset curation} We used satellite data from 2 different sources i.e. Commercial satellite dataset (private data repository) and LandSat8 satellite data (public data). Both the datasets have multispectral bands. The initial development and analysis were done using the data from commercial Pleiades-1B Satellite Sensor (AIRBUS Defence \& Space). It covers a few sections of Midland, Texas region. The spatial resolution of this data is $\sim 2m$ per pixel and 4 spectral bands. 3 of the spectral bands are in visible spectrum (Blue: $430-550\mu m$, Green: $490-610\mu m$, Red: $600-720\mu m$) and 1 in Near Infrared spectrum ($750-950\mu m$). All the data files are geo-tagged for each pixel. This dataset covered 17 frac pond sites with less than $10\%$ cloud cover. Sample from it dataset is shown in Figure~\ref{fig:sample_bodies}a.

To make our model more effective and useful, we used LandSat8 satellite -OLI~\cite{landsat8} to have a continuous stream of multispectral images from anywhere on the earth at a frequency of 2 weeks. LandSat8 orbits the earth every $99\; minutes$ in a sun-synchronous orbit at an altitude of $705\;kms$. LandSat8 acquires 740 scenes a day where each scene is $185 \times 180\;kms$. LandSat8 have 11 spectral bands information. The 3 bands in visible spectrum lies in the range $0.43-0.67\;\mu m$ with $30m$ resolution. The 4 bands in infrared regions lies in the range $0.85-1.38\;\mu m$ with $30m$ resolution. LandSat8 also have 2 thermal bands in wavelength range $10.6-12.51\;\mu m$ at $100m$ resolution. Along with these, LandSat8 data have a Panchromatic (PAN) band for visible spectrum ($0.50-0.68\;\mu m$) at resolution of $15m$. The data is captured by the sensor in 12-bit dynamic range. We downloaded data for 2 years (2018-2020), to do the initial development, analysis and evaluation of our $WQEI$. We manually downloaded the initial data from Midland and Permian basin in Texas. For later stage, we developed an end-to-end pipeline to pull data from LandSat8 and run $WQEI$ and create analysis report along with visualization

\begin{figure}[t]
\begin{center}
\includegraphics[width=0.7\linewidth]{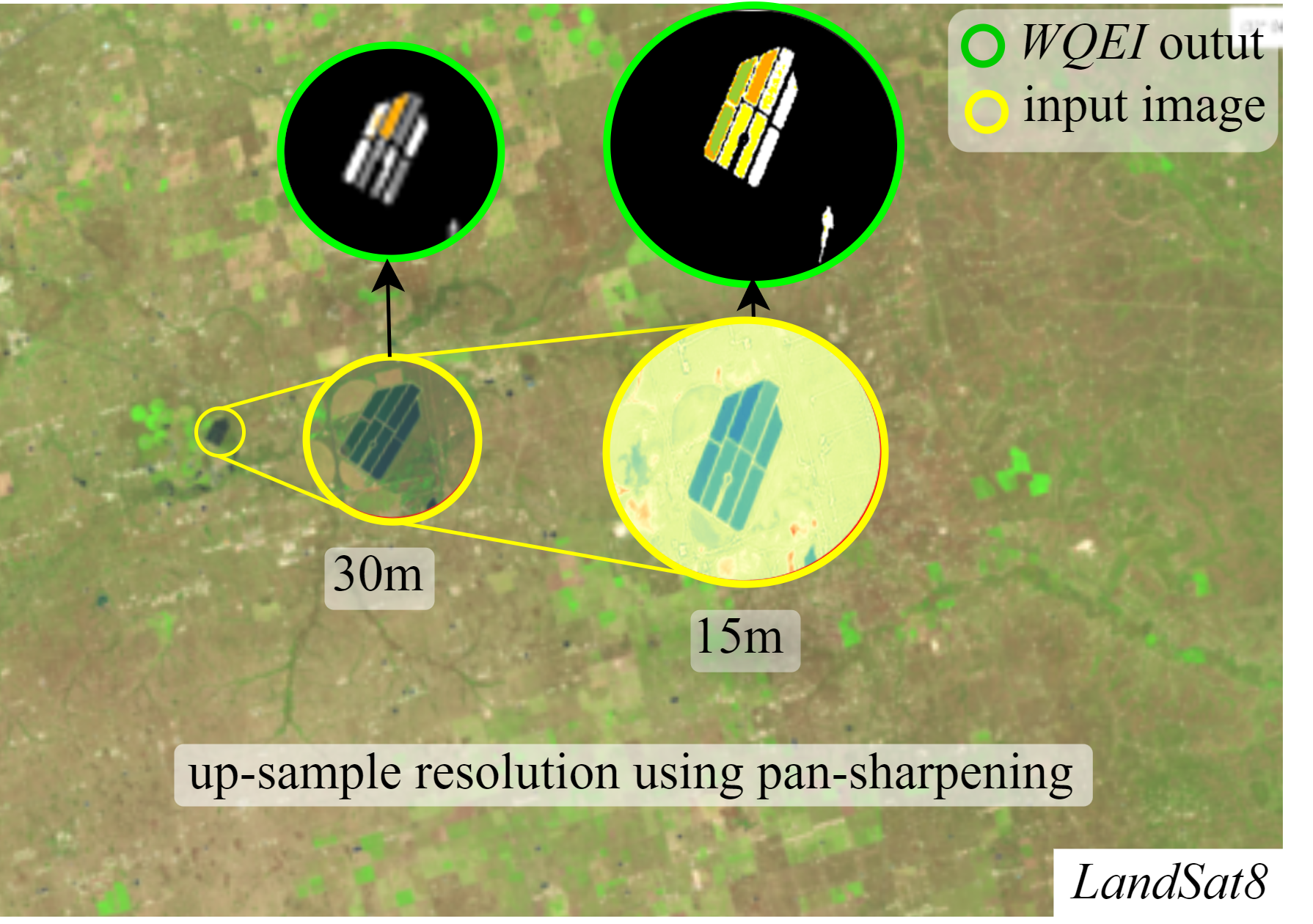}
\end{center}
\vspace{-0.45cm}
   \caption{The image shows the improvement in quality of $WQEI$ output by up-sampling the image using panchromatic band information. The shown sample is an irrigation pond.
   }
\label{fig:pan-sharp}
\end{figure}

\subsection{Pre-processing} \label{sec33}
\textbf{LandSat8 satellite data} is at $30m$ resolution, small ponds constitute very few pixels in the image. This way we does not have enough information to make a pixel level prediction. To overcome this issue, we used the panchromatic band data which is at higher resolution of $15m$ to improve the resolution of all visible and short infrared bands. First we normalize each band using mean ($\mu$) and standard deviation ($\sigma$) computed from the whole dataset downloaded. Then a high resolution image is created by up-sampling the visible bands using bi-linear interpolation method. The distribution ratio is then computed using \textit{Brovey} transform~\cite{brovey}. The decision for choosing \textit{Brovey} transform over others is motivated because of its simplicity and analysis by~\cite{vivone2014critical}. \textit{Brovey} transform is based on spectral modelling and increase the visual contrast in the high and low ends of the data's histogram. Each up-sampled band is multiplied by ratio of corresponding panchromatic pixel intensity to weighted sum of all multispectral bands~\cite{brovey}. 
\newline
\textbf{Lab reports} of water quality testing from frac ponds and other sources were organized to be in syn with satellite data. To minimize outliers, we selected only those sample which have atleast 20 test samples collected per pond over the span of 2 years. Since water sample collection for lab testing have different frequency, we shortlist only those were collected with 5 days window of LandSat8 satellite passing that location. We also normalized all the reading and unified the measuring scale of all the parameters measured in the lab reports (``pH", ``conductivity", ``temperature", ``algae", ``dissolved oxygen", etc).

\subsection{Water Quality Enhanced Index (WQEI)}
With the motivation insight mentioned in previous sections, we propose a novel Water Quality Enhanced Index. It is a new, invariant to spatial extent, robust analytical approach for characterizing the level of impurity spatially. We can map each and every section of the given water body to generate a spatial visual color-coded output representing the water quality. The $WQEI$ is developed in the following steps:

First a generic scan is done in the whole area, that scans for moisture content in the whole image, covers water body as well as soil around. Now we remove the potential confusers due to sand/soil/rocks or dust that settles on water temporarily, this is done by second term in equation~\ref{eq1}. The overall water impurity detector developed so far is as shown below
\begin{equation}
    imp\_detected = \frac{(R - NIR)}{R + NIR} - (\sqrt{B} + R )
    \label{eq1}
\end{equation}
where $R$ is red band ($0.65\mu m$), $G$ is green band ($0.55\mu m$), $B$ is expected value of (B$_1$ ($0.44\mu m$), B$_2$ ($0.48\mu m$)) i.e. blue band, $NIR$ is near infrared band ($0.86\mu m$). 
\newline
Next we observed that algae have a similar spectral signature to the index computed in equation~\ref{eq1}, and most of the algae growth is in or around the water body. We compute a similarity index with algae detection index, the intuition behind that is it will further remove the confuser element from the site. This increases the strength of the spectral signal representing water impurity. This step also ensures that algae growth on general vegetation around the water body or somewhere else in the area is removed. This is shown as below:
\begin{equation}
    amp\_sig = imp\_detected \; \circ \; \frac{G-R}{G+R}
    \label{eq2}
\end{equation}
where $imp\_detected$ is from equation~\ref{eq1}. Now we normalize this $amp\_signal$ with the sum of red and near infrared signal. This is a standard practice in literature as it scales the values in the range $-1$ to $+1$~\cite{normalized_idx}. Now to have a better fit according to the settings of the environment, we made the numerator and denominator differentiable function. 
We used regression-based approaches to estimate the values of variable parameters. The final estimated function is:
\begin{equation}
    WQEI_{1} = \frac{\{\frac{(R - NIR)}{R + NIR} - (\sqrt{B} + R )\} \; \circ \; \frac{G-R}{G+R}}{\sum_i^{h \times w} \alpha \times G - \beta \times (R + NIR)},
    \label{eq3}
\end{equation}
here $h$ and $w$ are the spatial dimension (height and width) of the image (water body) respectively, $\alpha$=2.74 and $\beta$=4.89 are the estimated values. Equation~\ref{eq3} is the final estimation index for water quality detection. 

\begin{figure}[t]
\begin{center}
\includegraphics[width=0.5\linewidth]{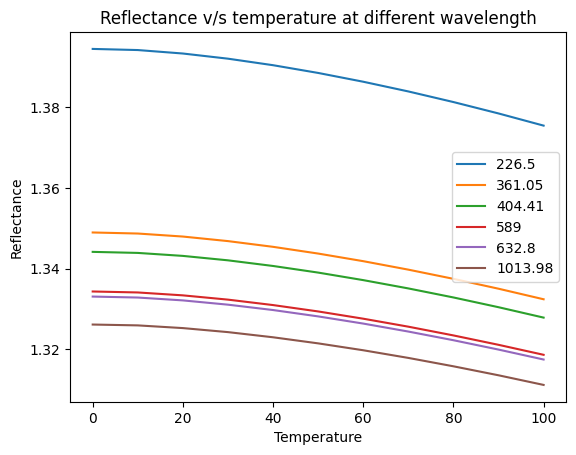}
\end{center}
\vspace{-0.45cm}
   \caption{This plot shows the dependence of reflectance of light at different wavelength versus temperature. The reflectance of water decreases are the temperature increases. Also, lower wavelengths have higher reflectance and wavelengths in near infrared region have lower reflectance
   }
\label{fig:sample_bodies}
\end{figure}

Next, we analyzed the water quality for the presence of chlorides, fluorides, and other salts. Our lab reports did not have direct tests for the presence of salts in the water. Salt level monitoring is also a very important factor when it comes to water quality. Specially used water pumped back into frac ponds from wells, or water discharged into natural playas for wildlife. Maintaining the level of salinity is very crucial in such cases. Our engineers performed tests for the conductivity of water samples collected from the ponds. Our basic assumption is here is that conductivity has a high correlation with the salinity of water~\cite{salinity,salinity2}.~\cite{bashkatov2003water, wezernak1976spectral} derived an approximate relation between the temperature of the water and its reflectance pattern. According to ~\cite{bashkatov2003water}, in the wavelength range $200\mu m-1000\mu m$ as the temperature of water increases, the reflectance of water reduces as shown in figure~\ref{fig:sample_bodies}. This study was done on clean water. A scientific report from NASA~\cite{wezernak1976spectral}, further backed this claim, on studying the reflectance pattern of different types of water bodies, e.g. salt water bodies, and clear water bodies at different temperatures. 

We further study the behavior of certain types of salt when they are mixed with water. Most of the saltwater mixture reactions as endothermic reactions~\cite{archer2002enthalpies}, which include, sodium chloride, phosphates, magnesium salts, etc. So we used the above two properties of water to propose a novel index to estimate the salinity in very small and shallow bodies of water. Most of the standard salinity indexes like normalized difference salinity index (NDSI) are not effective because they are designed to study the soil salinity or salinity of arid or semi-arid vegetation. The reflectance property of salts present in the soil is totally different from salts mixed in water. 

We start with the prior knowledge that water has a strong absorption towards red ($0.65\mu m$), near-infrared bands ($0.86\mu m$) and higher reflectance towards green ($0.55\mu m$), blue ($0.44\mu m$) bands~\cite{bashkatov2003water}. As the number of soluble impurities in water increases, its reflectance increase~\cite{malinowski2015detection}. There is a shift in the reflectance and absorption pattern of water towards red bands. The base salinity/impurity level detection of the water body is computed as:
\begin{equation}
    sal\_detect = (G + B) - \gamma (R + NIR),
    \label{eq4}
\end{equation}
where $R$ is red band ($0.65\mu m$), $G$ is green band ($0.55\mu m$), $B$ is expected value of (B$_1$ ($0.44\mu m$), B$_2$ ($0.48\mu m$)) i.e. blue band, $NIR$ is near infrared band ($0.86\mu m$) and $\gamma$ is a learnable coefficient. Next we explore the dependence on temperature, using the prior knowledge of~\cite{salinity,salinity2,bashkatov2003water,wezernak1976spectral} salt-water reaction being endothermic and temperate-reflectance relationship. Since temperature has a inverse relation with reflectance and variations in temperature are directly visible in the thermal bands ($10.8\mu m$) of LandSat8.
\begin{equation}
    sal\_detect = (G + B) - (\gamma (R + NIR))^{(\frac{\theta}{T})}
    \label{eq5}
\end{equation}
here $T$ is the thermal band ($10.8\mu m$) and $\theta$ is learnable coefficient.
This boost the weak signal from the from the small and shallow water pond as there is a shift towards the $R$ and $NIR$ bands with increase in impurity. Next we normalize equation~\ref{eq5}. The final Water Quality Enhanced Index is as shown below
\begin{equation}
    WQEI_2 = \frac{(G + B) - (\gamma (R + NIR))^{(\frac{\theta}{T})}}{NIR + R + G + B}
    \label{eq6}
\end{equation}
Equation~\ref{eq6} is the another estimated index for detection salinity of water. The final estimated/learnred values of $\gamma$=1.8 and $\theta$=2. The output ($WQEI_2$) of equation~\ref{eq6} is in range $-1$ to $+1$.

\section{Verification of Indexes}

\begin{figure*}[t]
\begin{center}
\includegraphics[width=\linewidth]{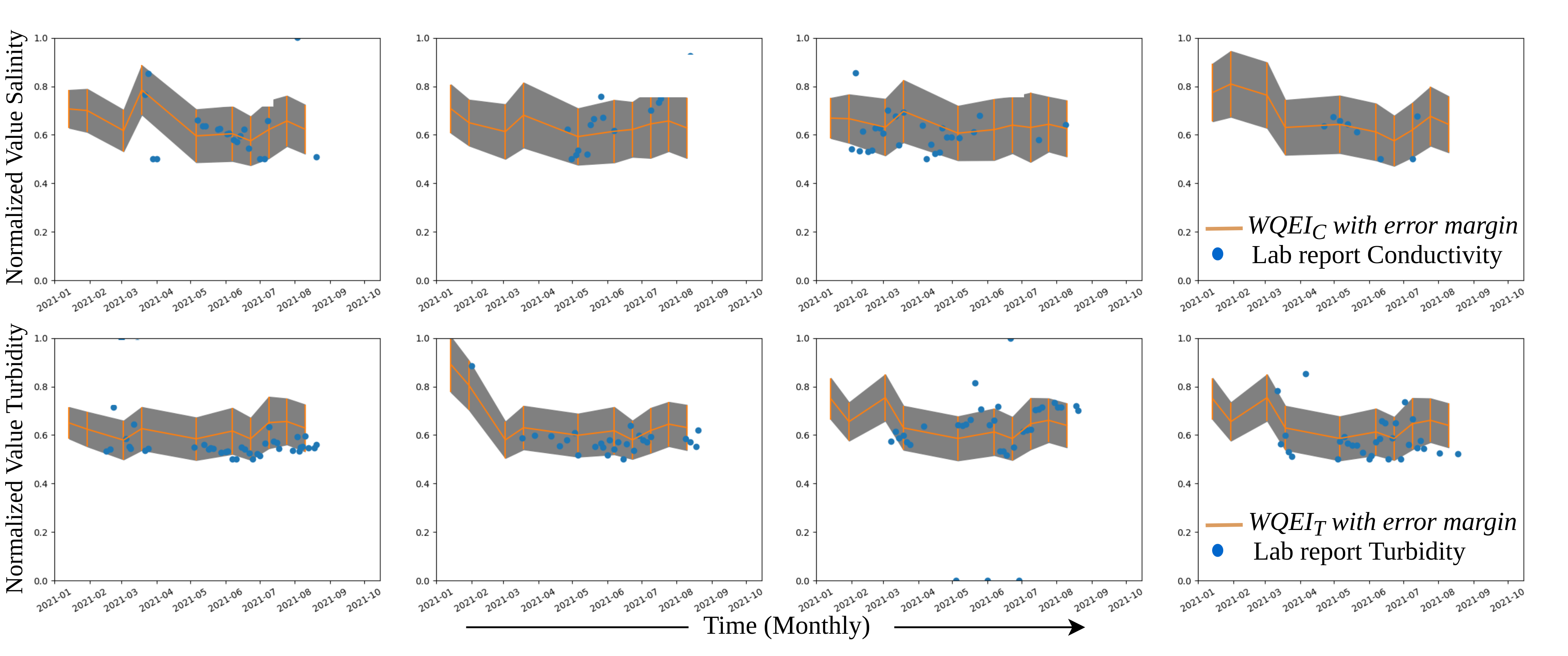}
\end{center}
\vspace{-0.45cm}
   \caption{Row 1 : Plot of $WQEI_C$ over 4 different ponds over the time period of one year, it shows the strong correlation of $WEQI_C$ with Lab reports for $Conductivity$. Row 2: Plot of $WQEI_T$ over 4 different ponds for 1 year and its strong correlation with Lab report of Conductivity computation.}
\label{fig:turb_conduct}
\end{figure*}

In this section, we will discuss the experiments that verified our created indexes and validation for our design choices. 
The validation tests are performed against the test done on water samples collected from the 49 frac ponds/irrigation ponds/playas. Here we will discuss in detail the types of tests and how we estimated the correlation between various tests and both the indexes created. 
\newline
\textbf{Satellites}: We used the multispectral images (remote sensing data) from two different satellites to ensure the generality of our methods. The multispectral data from commercial source has $\sim2m$ per pixel spatial resolution and 4 spectral bands, covering the visible and near-infrared regions. The spatial resolution is good enough for our small water ponds, we directly computed $WQEI_1$ and $WQEI_2$ for all the satellite images. In $11$ multispectral images, we have $17$ ponds sites. Since the output of each index is of the size of the input image ($H=20kms\times W=18kms$). The maximum size of any frac pond is $70m\times70m$, we cropped size of $100m\times100m$ (expected to cover the whole pond) around the $lat,\;long$ (approximate center of the pond) from the indexes output. Next, we compute the mean ($\mu^1_i, \mu^2_i$) and standard deviation ($\sigma^1_i, \sigma^2_i$) for each of the cropped outputs of indexes. 
\begin{equation}
    \mu^1_i, \sigma^1_i = \sum_{k=0}^{h\times w}P^1_k,\; \frac{1}{n}\sum_{k=0}^{h\times w}P^1_k - \mu^1_i  \;\;\forall \; (WQEI_1) 
\end{equation}
\begin{equation}
    \mu^2_i, \sigma^2_i = \sum_{k=0}^{h\times w}P^2_k,\; \frac{1}{n}\sum_{k=0}^{h\times w}P^2_k - \mu^2_i  \;\;\forall \; (WQEI_2) 
\end{equation}
\newline
where $\mu_i$ and $\sigma_i$ are mean and std of $i_{th}$ cropped output of indexes respectively, $P_k$ is the $k_{th}$ pixel in the cropped output.
\newline
\textbf{LandSat8} has spatial image resolution is $30m$ per pixel, we used panchromatic band information to up-sample the spatial resolution to $15m$ per pixel. Details about up-sampling transform are discussed in section~\ref{sec33}. On the up-sampled multispectral image, we compute both $WQEI_1$ and $WQEI_2$ for all the 49 ponds where each pond have approximately 18 images over the span of 2 years. 
Then we compute the mean ($\mu^1_i, \mu^2_i$) and standard deviation ($\sigma^1_i, \sigma^2_i$) of all images. In total (commercial satellite and LandSat8) we have approximately 1200 images taken at different time intervals.
\newline
\textbf{Lab test} reports data is cleaned and formatted to be compared directly with outputs ($WQEI_1,WQEI_2$) from satellite data. Each pond water sample is tested for the following parameters: $pH$, $Conductivity$, $Turbidity$, $Dissolved\; Oxygen$, $Temperature$, $H_2S$, $Depth$. To find out the relation of our indexes ($WQEI_1,WQEI_2$) with lab test parameters, we match each index with each of the test parameters recording using a matching criterion.

\begin{figure*}[t]
\begin{center}
\includegraphics[width=\linewidth]{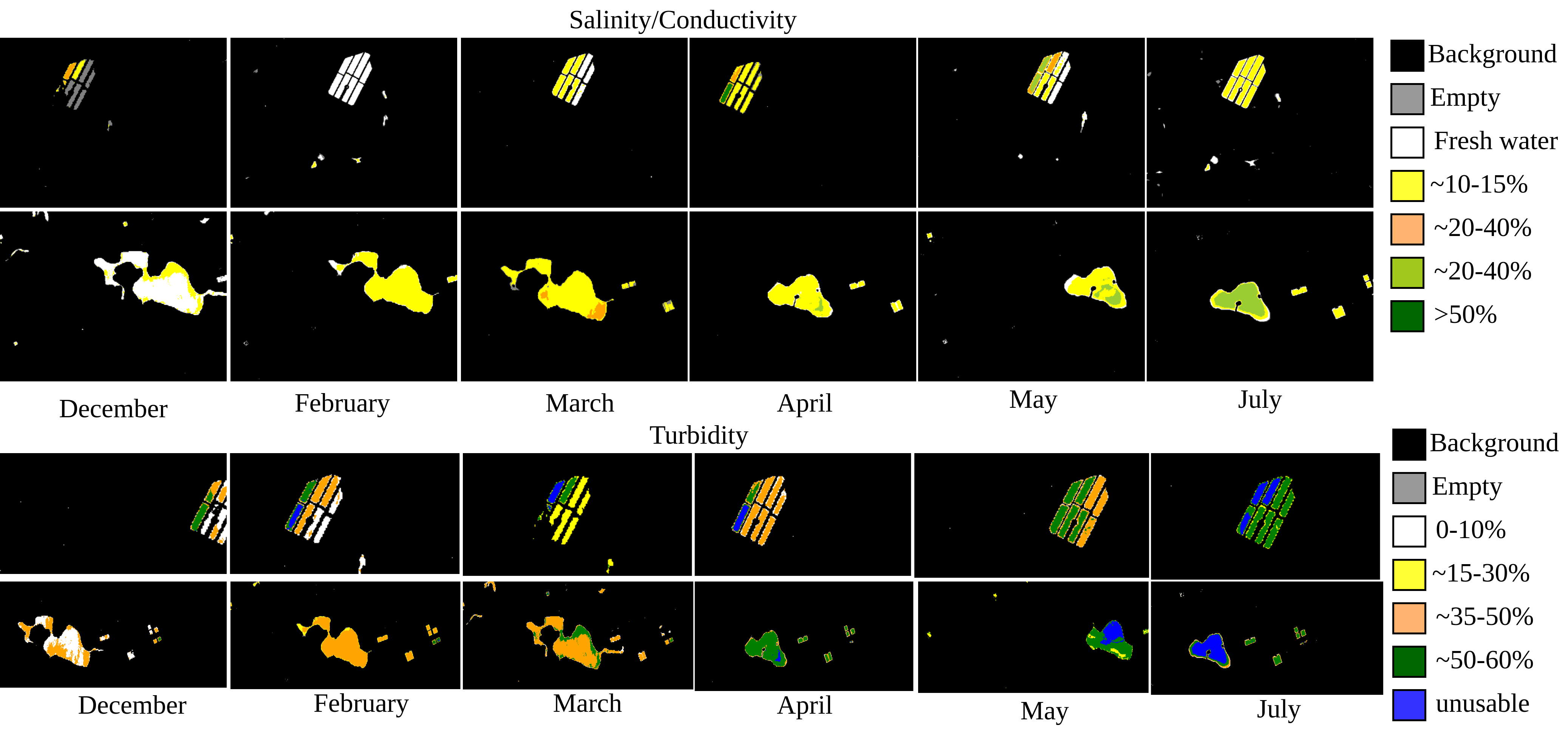}
\end{center}
\vspace{-0.45cm}
   \caption{Row 1\& 2: shows qualitative results of computing the $WQEI_C$ for a irrigation field and natural playas. The data is observed over 8 months and compared with Lab test also. Row 3 \& 4: shows qualitative results for computing the $WQEI_T$ for the same water bodies, and the data is observed over same time interval. We can see, towards the months of summer, the natural playa dries up and water is very saline and turbid, while the irrigation water pond was filled up from the near by canal in the month of july, showing less saline and more turbid, as turbidity increases temporarily as water is discharged in from canal}
\label{fig:turb_conduct}
\end{figure*}

\subsection{Matching Criterion:} The first matching criterion used our case is \textbf{Pearson Correlation} (PC). The reason for using it is that, Pearson Correlation coefficients are used in statistics to measure how strong a relationship is between two signals irrespective of the amplitude. It is defined as the covariance of the two signals divided by the product of their standard deviations. 
\begin{equation}
    \rho_{indexes, lab\_test} = \frac{cov (indexes, lab\_test)}{\sigma_{indexes}\; \circ \;\sigma_{lab\_test}},
\end{equation}
here $\sigma_{indexes}$ is $(\sigma^1_i, \sigma^2_i)$ and $\sigma_{lab\_test}$ is $(\sigma_{pH}, \sigma_{conductivity}, ..etc)$, $cov$ is covariance.
PC shows how significantly the shape of the curve plotted for indexes output readings ($\mu^1_i, \mu^1_i$) correlates well with  each lab test parameter ($pH, Conductivity,..etc$) over the time span of 2 years. For the PC, we are ignoring the magnitude because the signals (indexes v/s lab test) are from 2 totally different domain spaces. Magnitude optimization may lead to incorrect matching or no matching at all. So we started with curve shape matching first.

First we set all the learnable parameters ($\alpha, \beta, \gamma, \theta$) to 1 in equation~\ref{eq3} \&~\ref{eq6}. Next for each pond over the time span of 2 years ($\sim18$ readings), we compute PC for each index ($WQEI_1, WQEI_2$) with each lab test parameter ($pH, Conductivity, Turbidity, .. etc$). And picks the best match lab test parameter for each of our indexes. We found out that $WQEI_1$ has the highest correlation with $Turbidity$ and $WQEI_2$ has $Conductivity$.
\begin{equation}
    \begin{split}
    \begin{cases}
        \rho_1 = argmax(\textbf{PC}\;(WQEI_1,\; lab\_test))\\
        \rho_2 = argmax(\textbf{PC}\;(WQEI_2,\; lab\_test))
    \end{cases} \\
    \forall\;\; lab\_test \in (pH, Conductivity,..),
    \end{split}
\end{equation}
where \textbf{PC} is Pearson Correlation. We observed that $\rho_1$ points to $Turbidity$ and $\rho_2$ to $Conductivity$. Hence we found that our $WQEI_1$ is a measure of $Turbidity$ of water in the pond and $WQEI_2$ is a measure of $Conductivity$ of water estimated using satellite imagery. For the sake of simplicity now on, we will use $WQEI_T, WQEI_C$ for $WQEI_1$ and $WQEI_2$ respectively. Next we find the optimum value for out learnable coefficients ($\alpha, \beta, \gamma, \theta$) of indexes ($WQEI_T, WQEI_C$). For this we used second matching criterion using simple regression methods.

\textbf{Mean Square Error} (MSE) is used to estimate the right values of learnable coefficients. MSE is a model evaluation metric used for regression tasks. The main reason for using MSE as an evaluation/matching criterion is to estimate as precise as possible values for $Turbidity$ and $Conductivity$ of water in small ponds using satellite imagery. We used simple regression to estimate those values and optimized on minimizing the MSE for both the $WQEI_T$ and $WQEI_C$. The final indexes are as follows:
\begin{equation}
        WQEI_T = \frac{\{\frac{(R - NIR)}{R + NIR} - (\sqrt{B} + R )\} \; \circ \; \frac{G-R}{G+R}}{\sum_i^{h \times w} 2.74 \times G - 4.89 \times (R + NIR)},
    \label{eqWQEIT}
\end{equation}
\begin{equation}
    WQEI_C = \frac{(G + B) - (1.8 (R + NIR))^{(\frac{2}{T})}}{NIR + R + G + B}
    \label{eqWQEIC}
\end{equation}

With the help of our very simple and novel indexes, we can monitor the quality of water in small ponds, tanks, playas, etc using satellite imagery in the remotest section of the world. We can reuse the wastewater to recharge the natural playas. We can point out the location where the water is unfit for drinking by the wildlife, provide alternate sources and block the unfit pond, or tank.

\begin{table}
\parbox{.45\linewidth}{
\centering
\begin{tabular}{cccccc}

\multicolumn{1}{c}{} & \multicolumn{5}{c}{\textit{Ponds (MSE) - Turbidity}} \\
\multicolumn{1}{c}{\multirow{-2}{*}{\textit{Methods}}} & \cellcolor[HTML]{EFEFEF}\textit{P1} & \cellcolor[HTML]{EFEFEF}\textit{P2} & \cellcolor[HTML]{EFEFEF}\textit{P3} & \cellcolor[HTML]{EFEFEF}\textit{P4} & \cellcolor[HTML]{EFEFEF}\textit{P5} \\
\textit{TI1}~\cite{saylam2017assessment} & 0.955 & 0.880 & 0.861 & 0.797 & 0.862 \\
\rowcolor[HTML]{EFEFEF} 
\textit{TD2}~\cite{yarger1973water} & 0.768 & 0.724 & 0.736 & 0.777 & 0.779 \\
\textit{MLR}~\cite{liu2019modelling} & 0.645 & 0.655 & 0.588 & 0.511 & 0.680 \\
\rowcolor[HTML]{EFEFEF} 
\textit{GEP}~\cite{liu2019modelling} & 0.563 & 0.541 & 0.581 & 0.410 & 0.471 \\
\textit{\begin{tabular}[c]{@{}l@{}}Reg\\ Model~\cite{hossain2021remote}\end{tabular}} & 0.157 & 0.243 & 0.159 & 0.151 & 0.140 \\
\rowcolor[HTML]{EFEFEF} 
\textit{\begin{tabular}[c]{@{}l@{}}Neural\\ Net (Ours)\end{tabular}} & 0.218 & 0.245 & 0.273 & 0.395 & 0.206 \\
\textit{\textbf{WQEI$_T$}} & \textbf{0.097} & \textbf{0.028} & \textbf{0.017} & \textbf{0.040} & \textbf{0.035}
\end{tabular}
\vspace{0.1cm}
\caption{Comparison of $WQEI_T$ with the popular methods of detecting water turbidity. $WQEI_T$ have lower MSE than other methods when tested on 5 randomly selected ponds.}
}
\hfill
\parbox{.45\linewidth}{
\centering
\begin{tabular}{cccccc}
 & \multicolumn{5}{c}{\textit{Ponds (MSE) - Salinity}} \\
\multirow{-2}{*}{\textit{Methods}} & \cellcolor[HTML]{EFEFEF}\textit{P1} & \cellcolor[HTML]{EFEFEF}\textit{P2} & \cellcolor[HTML]{EFEFEF}\textit{P3} & \cellcolor[HTML]{EFEFEF}\textit{P4} & \cellcolor[HTML]{EFEFEF}\textit{P5} \\

\textit{SI4}~\cite{khan2005assessment} & 0.757 & 0.997 & 0.892 & 0.704 & 0.932 \\
\rowcolor[HTML]{EFEFEF} 
\textit{SI6}~\cite{bannari2008characterization} & 0.799 & 0.655 & 0.618 & 0.890 & 0.860 \\
\rowcolor[HTML]{FFFFFF} 
\textit{SI7}~\cite{jamalabad2004forest} & 0.508 & 0.648 & 0.669 & 0.669 & 0.550 \\
\rowcolor[HTML]{EFEFEF} 
\textit{SAVI}~\cite{alhammadi2008detecting} & 0.444 & 0.545 & 0.497 & 0.589 & 0.633 \\
\rowcolor[HTML]{FFFFFF} 
\textit{NDSI}~\cite{noureddine2014new} & 0.358 & 0.466 & 0.341 & 0.438 & 0.353 \\
\rowcolor[HTML]{EFEFEF} 
\textit{VSSI}~\cite{mousavi2017digital}& 0.165 & 0.179 & 0.177 & 0.165 & 0.177 \\
\rowcolor[HTML]{FFFFFF} 
\textit{MSAVI}~\cite{khan2001mapping} & 0.106 & 0.175 & 0.298 & 0.217 & 0.131 \\ 
\rowcolor[HTML]{EFEFEF}
\textit{\begin{tabular}[c]{@{}c@{}}Neural \\ Net (Ours)\end{tabular}} & 0.254 & 0.137 & 0.141 & 0.196 & 0.256 \\
\rowcolor[HTML]{FFFFFF} 
\textit{\textbf{WQEI$_C$}} & \textbf{0.082} & \textbf{0.068} & \textbf{0.099} & \textbf{0.063} & \textbf{0.081}
\end{tabular}
\vspace{0.1cm}
\caption{Comparison of $WQEI_C$ to detect salinity with the most popular indexes that exists in literature. It can be seen $WQEI_C$ have lower mean square error as compared with other methods. We performed it on 5 randomly selected ponds and computed the salinity using each index listed from satellite images and the corresponding lab report.}
}
\end{table}


\vspace{-0.35cm}
\subsection{Estimation with Neural Networks:} In this section we will discuss the neural network architecture we designed to estimate $Turbidity$ and $Conductivity$. We started with a small VGG16~\cite{simonyan2014very} network pre-trained on natural images dataset (imageNet)~\cite{deng2009imagenet}. We used convolutional layers only of the VGG16 as feature extractor, appended 2 Fully Connected (FC) layers initialized randomly. The final layer has 2 output neurons. The training data used is 1200 satellite images cropped out for each pond at different timestamps. The ground truth are the lab test reports discussed in section~\ref{sec33}. We created a 80-10-10\% train-val-test split of the dataset. 
\newline
\textbf{Loss function} used is inspired from~\cite{chai2014root,kumar2021stressnet}. Since we are directly predicting a value for $Turbidity$ and $Conductivity$, the output of last layer is passed through a softmax function. The loss function used is simple MSE for optimization. Each input image is of size $100\times100$ with 12 bit depth information per pixel. The batch size is 32 images with a learning rate of 0.005 at starting and reduces by a factor of 0.5 after every 10 epochs. Stochastic Gradient Descent is used as optimizer for the network. The network is trained for 30 epochs.

\section{Results and Conclusions}

In this study, we utilize multi-spectral data (LandSat8) to monitor water bodies that are dispersed in a large region. By creating the Water Quality Enhanced Index ($WQEI_T, WQEI_C$), we show:
\begin{itemize}
    \item Comparing our $WQEI_T, WQEI_C$ model prediction with laboratory water quality measurements, our results show that applying our model can achieve frequent water body monitoring with sufficient accuracy.
    \item  The time series plot (Figure~\ref{fig:turb_conduct}) shows that $WQEI$ is a good indicator of water quality.  More specifically, $WQEI_C$ shows a strong correlation with water conductivity and $WQEI_T$ with water turbidity.
    \item  Compared with traditional water quality index models, the $WQEI$ formulation includes a water impurity term (Equation~\ref{eq1}, a signal amplification term (Equation~\ref{eq2}) and a normalization term (Equation~\ref{eq3}). The combination of multiple terms we introduced leads to better performance, especially for small-sized water bodies.
\end{itemize}
The proposed methods are evaluated on the criterion of Mean Squared Error (MSE). They are also compared to standard indices for detecting $Turbidity$ and $Conductivity/Salinity$. As shown in table 1\&2, $WQEI_T$ and $WQEI_C$ performs better than most existing methods~\cite{khan2005assessment,noureddine2014new,bannari2008characterization,jamalabad2004forest,alhammadi2008detecting,khan2001mapping,mousavi2017digital} \&~\cite{saylam2017assessment,yarger1973water, liu2019modelling, hossain2021remote}. Even with limited data, our proposed neural network accurately detects water salinity and turbidity. Further research is needed to stabilize these results: the neural network proposed here primarily functions as validation of the feasibility of this idea.

\subsection{Industrial Deployment} We deployed a fully function model of our detection methods in industrial application for monitoring and maintaining water quality at independent oil and gas organization. We will release the complete source code and details in the final publication and plan to present these findings to several oil and gas trade associations.

\section{APPENDIX A: Additional Experimental Results Visualizations}\label{sec:additional_exps}

\begin{figure}[h]
\begin{center}
\includegraphics[width=0.8\linewidth]{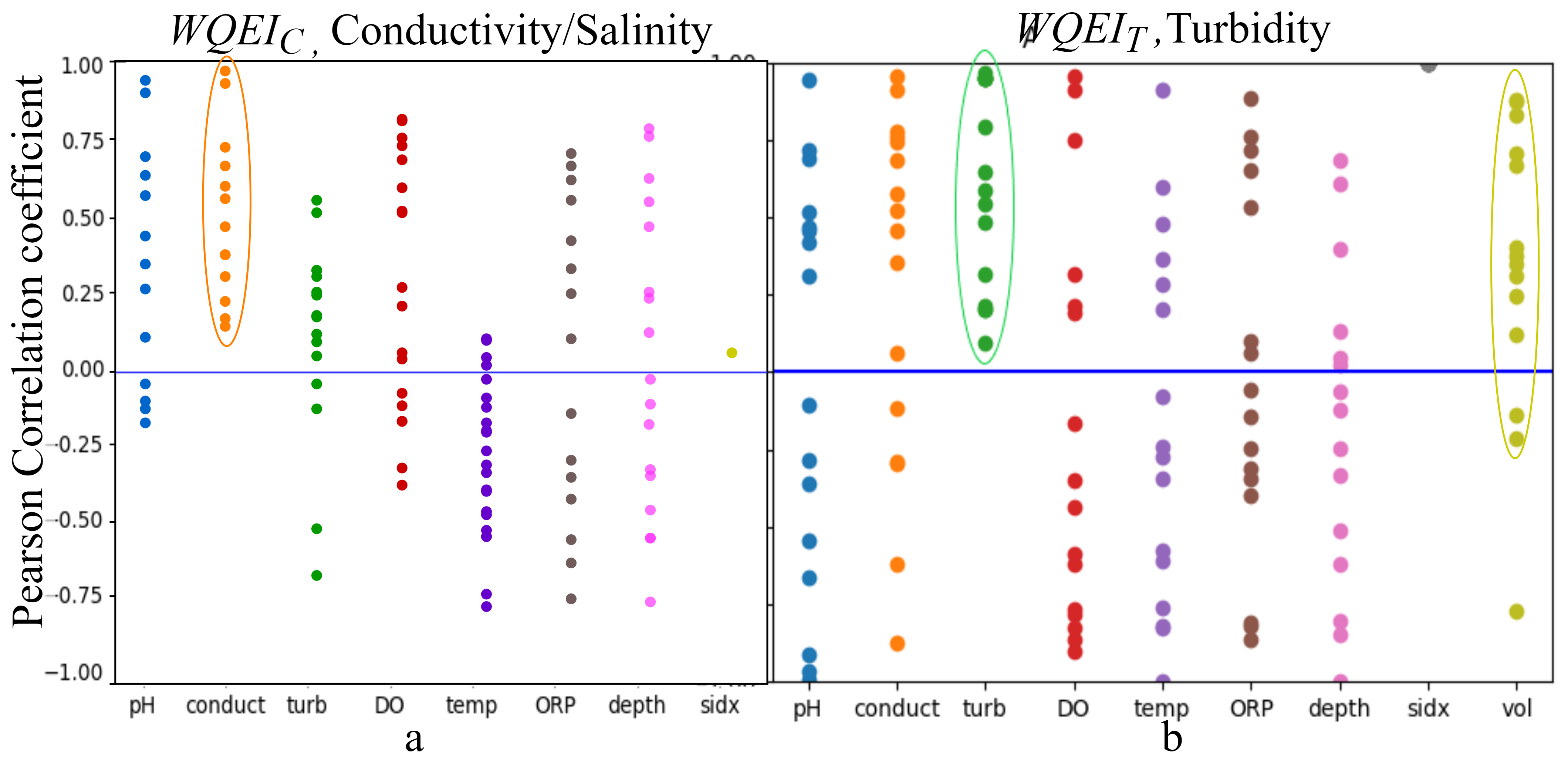}
\end{center}
   \caption{Pearson correlation plot of $WQEI_C$ \& $WQEI_T$ with lab test parameters. The oval marks shows the strongest correlation.}
\label{fig:pcp}
\end{figure}
\subsection{Pearson Correlation plots}
As discussed in section \textcolor{red}{4.1} in the main paper, the correlation plots for for equation \textcolor{red}{10} are shown in figure~\ref{fig:pcp}. Figure~\ref{fig:pcp}a shows the Pearson correlation coefficient plot for $WQEI_C$ with each of the lab test parameter ($pH, Conductivity, Turbidity, etc$). As shown with orange oval mark in figure~\ref{fig:pcp}a, $Conductivity$ has highest correlation with $WQET_C$. Similary figure~\ref{fig:pcp}b shows the Pearson correlation coefficient plot for $WQEI_T$ with each of the lab test parameter ($pH, Conductivity, Turbidity, etc$). The green oval mark shows a strong correlation of $WQEI_T$ with Turbidity. We also observed that it has significant correlation with volumne of water too in the pond. This way, using $WQEI_T$, we can make a relative estimate of volumne of water in the water body using satellite imagery only.


\subsection{Qualitative results}
\begin{figure*}[h]
\begin{center}
\includegraphics[width=0.8\linewidth]{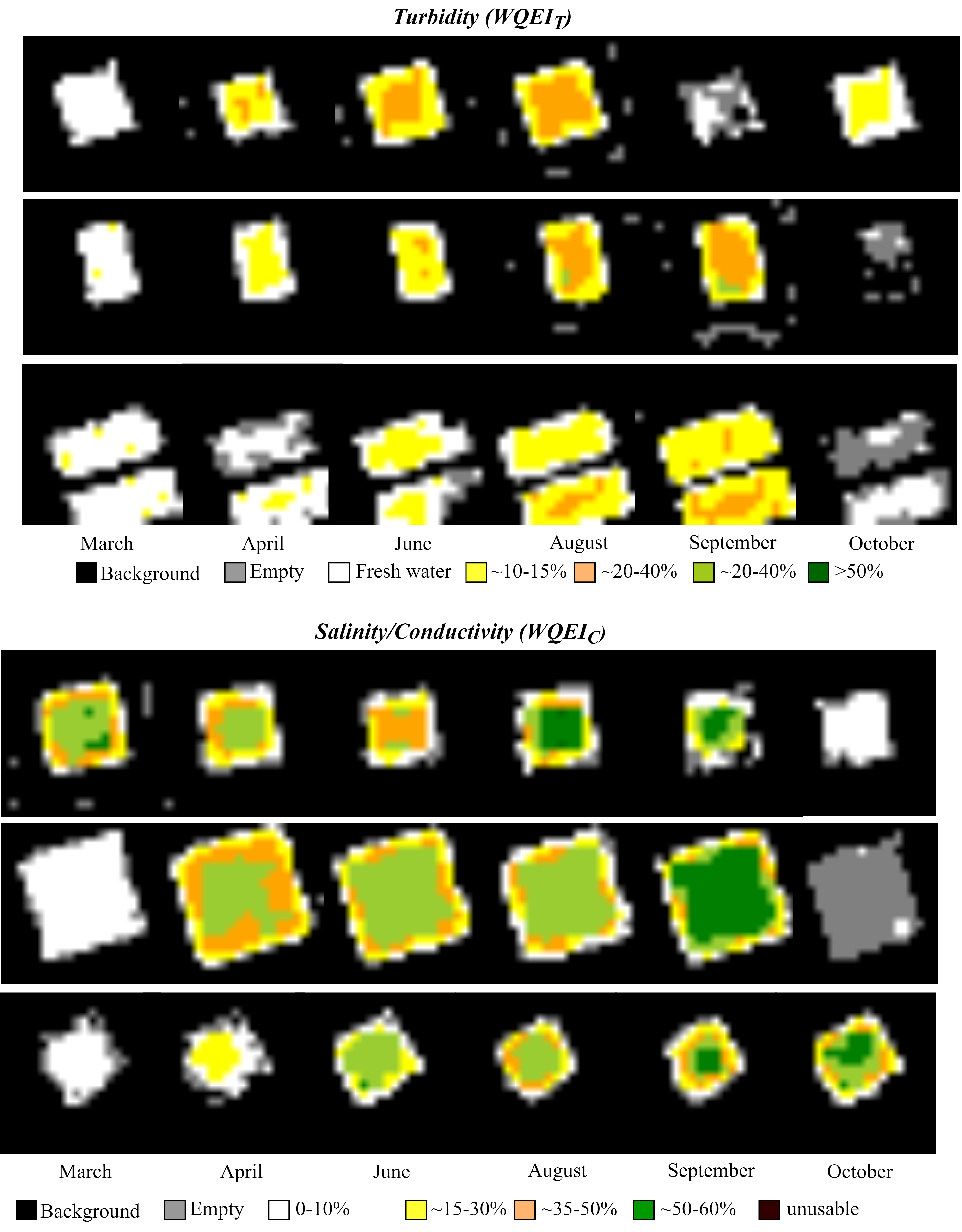}
\end{center}
   \caption{Qualitative results on frac ponds from LandSat8 satellite for turbidity and salinity detection using $WQEI_T$ \& $WQEI_C$ respectively. The results are shown from the start of summer in midland, Texas region until start of winter.}
\label{fig:q_res}
\end{figure*}

Figure~\ref{fig:q_res} shows some of the qualitative results on very small frac ponds in Midland, Texas region. The indexes are computed on images from LandSat8 satellite. We picked a mixture of ponds, the ones from which, water is used very frequently and refilled frequently and the ones from which, water is used only once in 3-4 months. Row 1 and 2 (Turbidity ($WQEI_T$)) shows the ponds which are used very frequently (almost every 3 weeks) and refilled very frequently. We can see, most of the time the water stays turbid.
Row 3 (Turbidity ($WQEI_T$)) shows very less used pond (3-4 months, the water stays calm and idle but the pond starts to dry up over time. We can see for the month of march, april, june of row 3, the turbidity is low and the pond starts to dry up, refilled back in next months, based on usage, we can see the increase in turbidity.

\begin{figure}[t]
\begin{center}
\includegraphics[width=0.7\linewidth]{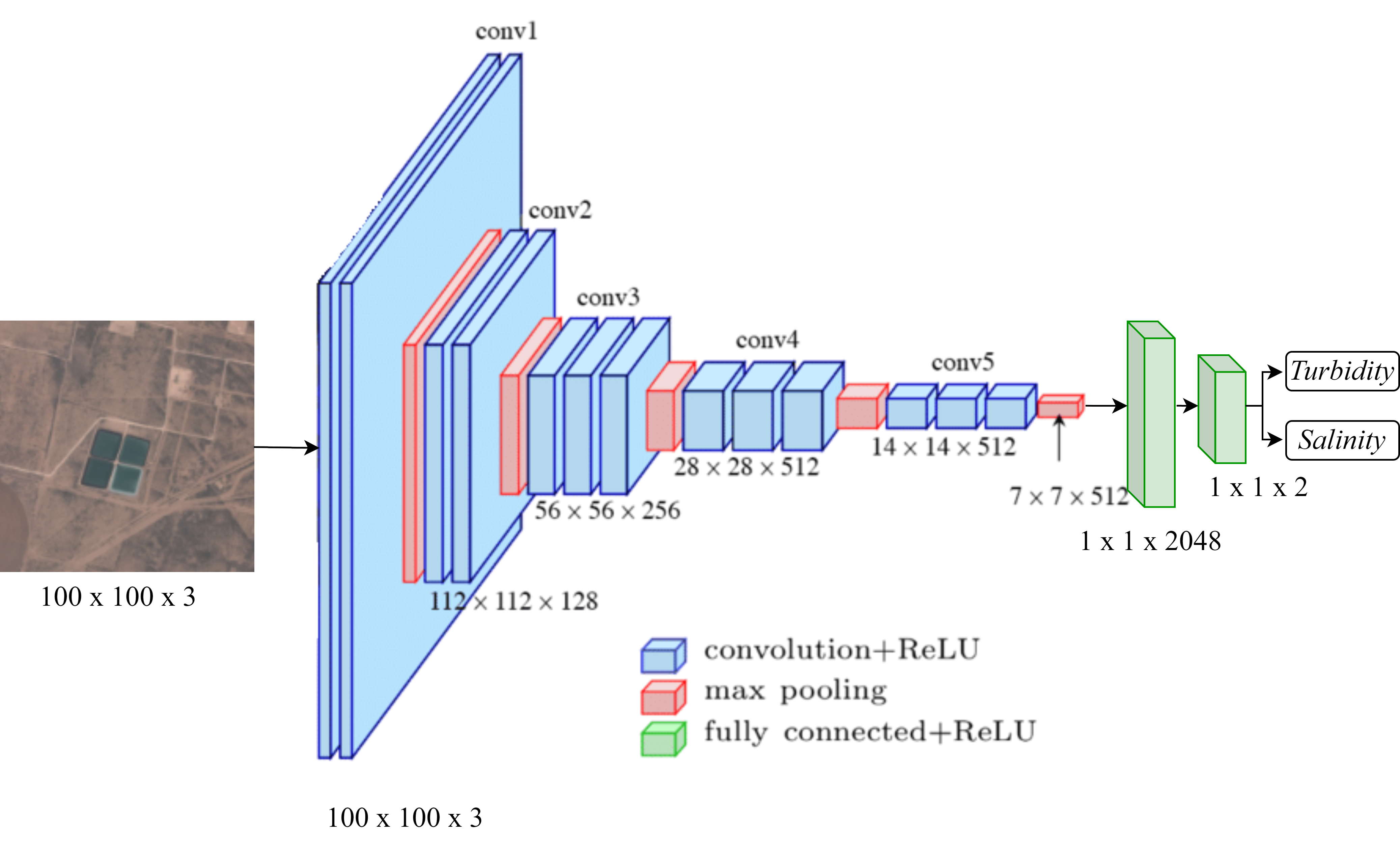}
\end{center}
   \caption{Neural Network architecture used for detecting water turbidity and conductivity/salinity.}
\label{fig:vgg}
\end{figure}

Row 4,5,6 (Salinity/Conductivity ($WQEI_C$)) shows how the salinity of the frac pond changes over time. We observed that the salinity is very high in the months towards the end of summer, as the fracturing process is at higher rate and a lot of water is pumped back from the well back to the pond. Salinity is very high also in those cases when the water in the pond has almost dried up. As we can see in row 5 in month is September (this pond was unused), towards the end of summer, almost all the water in the pond dries up and we can see high concentration of salts in the ponds before it dries up completely.

\subsection{Neural Network Architecture}
As mentioned in the main paper we tested a simple VGG16~\cite{simonyan2014very}. Here we will add more details about the architecture and the observations on training a neural network on remote sensing data for our problem set. The input image size is $100\times100\times3$. We used pre-trained VGG16(pre-trained on imagenet dataset~\cite{deng2009imagenet}) for input image feature extraction to a feature map of size $7\times7\times512$. Fully connected layer with ReLu activation are initialized randomly. The output is passed through a softmax function to predict score of turbidity and salinity/conductivity. The architecture diagram is shown in Figure~\ref{fig:vgg}.

\textbf{Limitations:} The training data used is very limited in our case and the predictions are only the score of $Turbidity$ and $Conductivity/Salinity$ of the whole pond. Limited data problem arises in almost all the fields when it comes to remote sensing. As the terrain covered to do any kind of anomaly detection (e.g. water quality, sand quality, GHG emissions, etc) is extremely large (~1000s of miles) and it is very challenging to generate ground truth information in such cases. For example in our case only, the data from 49 ponds only was a major challenge, our team spent 2 years of time and millions of dollars just to monitor such a small number of ponds. Unlike natural images object detection where simple objects (e.g. car, truck, animal, person) seen in surroundings can be annotated with very minimal amount of knowledge. We need specialized equipments and expertise in the subject matter (e.g. chemical engineering, hydrologist, geologist, etc) to check presence of certain types of impurities. And after that the next challenge is understanding of multispectral data.

\textbf{Index creation}~\cite{khan2005assessment,noureddine2014new,bannari2008characterization,jamalabad2004forest,alhammadi2008detecting,khan2001mapping,mousavi2017digital,saylam2017assessment,yarger1973water, liu2019modelling, hossain2021remote} is one way where people have done analysis of multispectral data from satellites using remote sensing. Index creation needs expertise in both the multispectral imagery from satellite and chemical properties of object of interest. Indexes are very much prone to change in environmental conditions and the change in terrain. The simplistic neural network used for the problem gives green light to application of machine learning in remote sensing. We can predict different kind of impurities score by just using one neural network, instead of using a specialized index for each one. Our future work in this domain is to work towards using an encoder-decoder network architecture, so that we can generate a spatial map of impurities in the whole water body instead of a single score. But that is also going to need lab test on water samples collected from different sections of same water body.


\bibliography{report} 
\bibliographystyle{spiebib} 

\end{document}